\pdfoutput=1

\documentclass[11pt]{article}
\usepackage{acl}

\usepackage{times}
\usepackage{amsmath}
\usepackage{latexsym}
\usepackage{pdfpages}
\usepackage{times}
\usepackage{amsmath}
\usepackage{makecell}
\usepackage{latexsym}
\usepackage{indentfirst}
\usepackage{enumerate}
\usepackage{multirow}
\usepackage{graphicx}
\usepackage{acl}
\usepackage{color}
\usepackage{arydshln}
\usepackage[T1]{fontenc}

\usepackage[utf8]{inputenc}

\usepackage{microtype}
\usepackage{graphicx}
\title{SCL-RAI: Span-based Contrastive Learning with Retrieval Augmented Inference for Unlabeled Entity Problem in NER}

\author{Shuzheng Si$^{1,2}$\footnotemark[1] , Shuang Zeng$^{1,2}$\footnotemark[1] , Jiaxing Lin$^{1,2}$, Baobao Chang$^{1}$\footnotemark[2] \\
$^1$Key Laboratory of Computational Linguistics, Peking University, MOE, China \\ 
$^2$School of Software and Microelectronics, Peking University, China \\
\texttt{\{sishuzheng,jxlin\}@stu.pku.edu.cn \{zengs,chbb\}@pku.edu.cn}}

\begin{document}
\maketitle

\begin{abstract}
Named Entity Recognition is the task to locate and classify the entities in the text. 
However, \textit{Unlabeled Entity Problem} in NER datasets seriously hinders the improvement of NER performance.
% ...
This paper proposes \textbf{SCL-RAI} to cope with this problem.
Firstly, we decrease the distance of span representations with the same label while increasing it for different ones via span-based contrastive learning, which relieves the ambiguity among entities and improves the robustness of the model over unlabeled entities. 
Then we propose retrieval augmented inference to mitigate the decision boundary shifting problem.
Our method significantly outperforms the previous SOTA method by 4.21\% and 8.64\% F1-score on two real-world datasets.

\end{abstract}

\renewcommand{\thefootnote}{\fnsymbol{footnote}}
\footnotetext[1]{Equal contribution.}
\footnotetext[2]{Corresponding author.}
\renewcommand{\thefootnote}{\arabic{footnote}}

\section{Introduction}
\begin{figure}
    \centering
    \includegraphics[width=7.6cm]{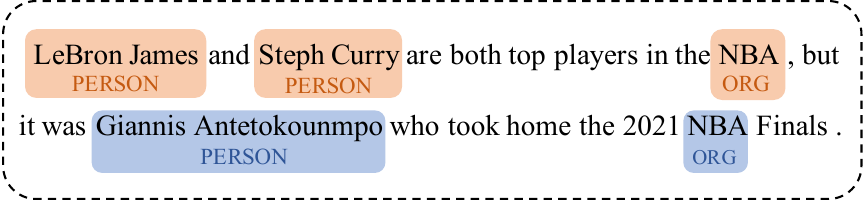}
    \caption{A toy case to show \textit{Unlabeled Entity Problem}. 
    The entities ``LeBron James'' , ``Steph Curry'' and the first ``NBA'' are correctly labeled while
    ``Giannis Antetokounmpo" and the second ``NBA'' are unlabeled.}
    \label{fig_example}
\end{figure}

\noindent
As a fundamental task in NLP, Named Entity Recognition aims to locate and classify named entities in the text.
Due to the large-scale well-annotated datasets, deep-learning based methods \citep{li2022unified,DBLP:conf/naacl/DevlinCLT19} have achieved great success. 
However, in real-world datasets, such as \citet{DBLP:conf/aaai/LingW12} with 112 fine-grained named entity tags, a large set of entity classes may cause inevitable missing annotations. 
Moreover, to obtain large NER datasets in practical scenarios, the distant supervision approach \citep{DBLP:conf/kdd/RenEWTVH15,DBLP:journals/corr/Fries0RR17} may make this problem even worse, since the entity dictionary cannot cover all entities. 
Previous work \citep{DBLP:conf/iclr/LiL021, DBLP:journals/tkde/ShangLJRVH18} find that this problem seriously hinders the performance of the NER model and name this problem as \textit{Unlabeled Entity Problem}. As shown in Figure \ref{fig_example}, the unlabeled second ``NBA'' may confuse model and introduce unnecessary noise.

\begin{figure}
    \centering
    \includegraphics[width=8cm]{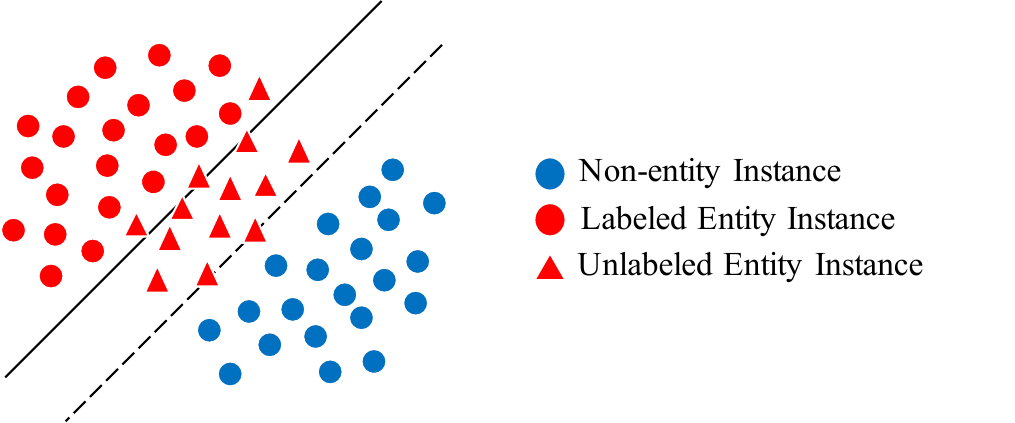}
    \caption{Illustration for decision boundary shifting phenomenon. The solid line is the learned boundary from datasets with unlabeled entities, the dashed line represents the expected boundary.}
    \label{decision boundary bias phenomenon}
\end{figure}

To cope with this problem, several attempts from different perspectives have been proposed.
Inspired by positive-unlabeled (PU) learning \cite{DBLP:conf/ecml/LiL05},  \citet{DBLP:conf/acl/PengXZFH19} use a weighted loss to assign low weights to false negative words and build distinct binary classifiers for different entity types. 
However, they require prior information or heuristics \cite{li-etal-2022-rethinking} and the unlabeled entities still misguide the classifiers, bringing ambiguity among neighboring entities \cite{DBLP:conf/iclr/LiL021}.
\citet{DBLP:conf/coling/YangCLHZ18, jie-etal-2019-better} introduce the Partial CRF \cite{DBLP:conf/icml/LaffertyMP01} to marginalize the instances that are consistent with the incomplete annotation.
However, they require additional well-annotated corpus to get ground truth negative instances, which are usually unavailable in practice.
Recently, \citet{DBLP:conf/iclr/LiL021} 
perform down-sampling among non-entity instances within annotation when computing loss function, in order to mitigate the misguidance from possible unlabeled entities. 
\citet{li-etal-2022-rethinking} further propose a weighted and adaptive sampling distribution to introduce direction to real unlabeled entities when down-sampling.
However, the inherent randomness of sampling strategy may still keep some unlabeled entities when computing loss then make the decision boundary biased \citep{li2022who}. 
As shown in Figure \ref{decision boundary bias phenomenon}, the learned decision boundary for training example containing unlabeled entity instances tends to shift from the expected boundary towards the entity side.
The previous works do not consider this problem.


To deal with these challenges, 
this paper proposes the \textbf{S}pan-based \textbf{C}ontrastive \textbf{L}earning with \textbf{R}etrieval \textbf{A}ugmented \textbf{I}nference (\textbf{SCL-RAI}) to tackle \textit{Unlabeled Entity Problem}, which mitigates the limitations mentioned above, i.e., demanding additional corpus, ambiguity among neighboring entities and decision boundary shifting problem.
Firstly, SCL-RAI tries to decrease the distance among span representations with the same labels while increasing it for different ones.
Benefiting from our well-designed span-based contrastive learning, the ambiguity between entities is mitigated by the increased representation distance, so the model can capture the differences among different entity labels. We show in experiment that this contrastive learning objective also improves the model robustness under unlabeled entities. Furthermore, we propose Retrieval Augmented Inference to relieve decision boundary shifting phenomenon. 
It caches the center point representation for each entity type from the training set. 
Then, it computes a label distribution via cached representation and interpolates it with the distribution from the backbone NER model.
Experiments on two real-world datasets show that SCL-RAI significantly outperforms previous SOTA methods. 


\section{Methodology}
\noindent
Our SCL-RAI consists of three modules: Span-based NER Model, Span-based Contrastive Learning, and Retrieval Augmented Inference.

\begin{figure*}[ht]
\centering 
\includegraphics[width=1.0\linewidth]{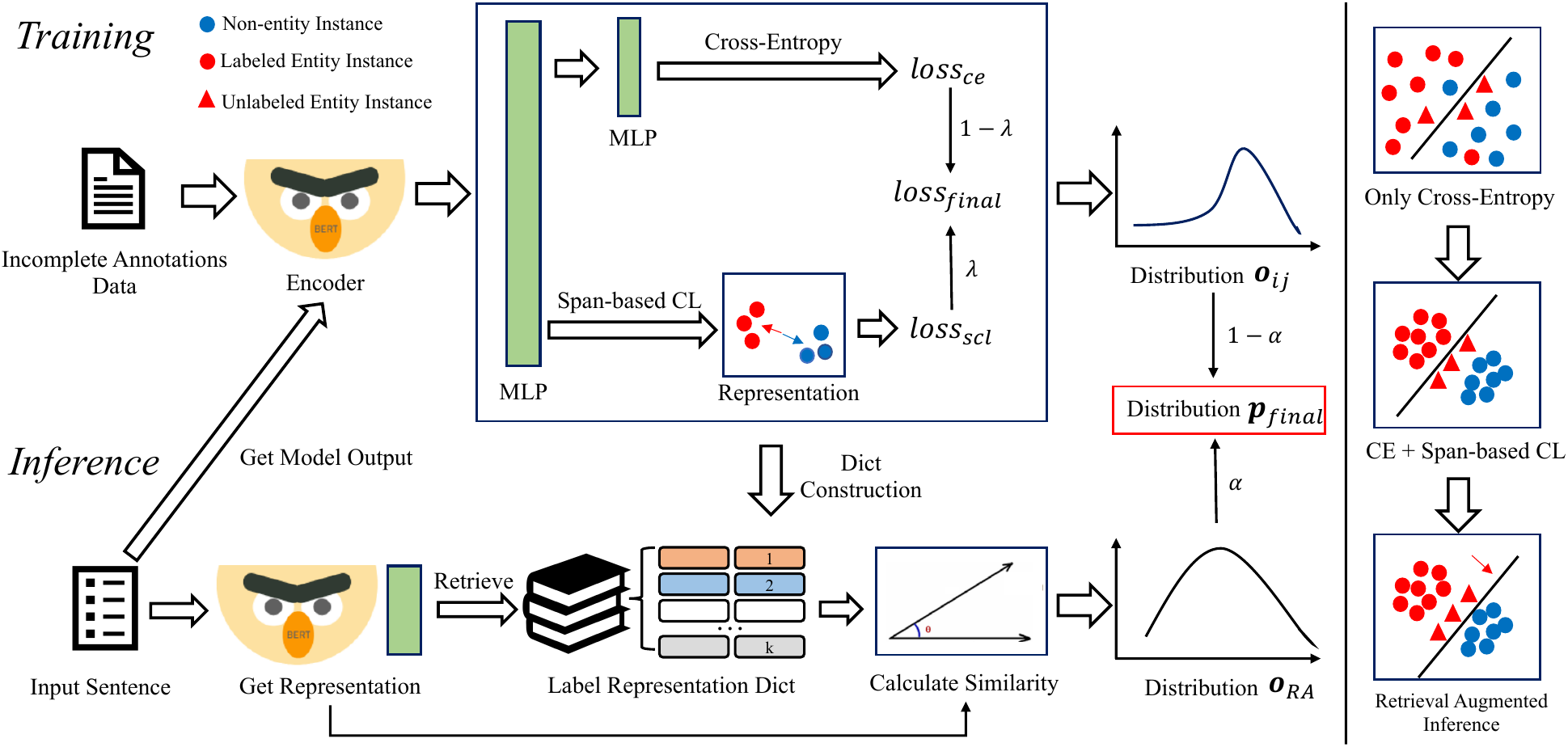} 
\caption{General architecture of SCL-RAI.
} 
\label{Fig_new} 
\end{figure*}

\subsection{Span-based NER Model}
\noindent
Span-based NER models have shown a strong ability to solve NER task, especially in flat NER and nested NER problem \citep{yu-etal-2020-named}. 
For fair comparison, we follow \citet{DBLP:conf/iclr/LiL021,li-etal-2022-rethinking} on the design of Span-based NER model.
Firstly, we use BERT \cite{DBLP:conf/naacl/DevlinCLT19} as the text encoder to get the representations for words in sentence $x$:
\begin{align}
[\textbf{h}_1,\textbf{h}_2,...,\textbf{h}_n] = BERT(x)
\end{align}
where $h_i$ is the representation for word $x_i$. For each text span $s_{i,j}$ ranging from $i$-th word and $j$-th word in $x$, we get the span representation $\textbf{s}_{i,j}$ as:
\begin{align}
\textbf{s}_{i,j} = \textbf{h}_i \oplus \textbf{h}_j \oplus (\textbf{h}_i - \textbf{h}_j) \oplus (\textbf{h}_i \odot \textbf{h}_j)
\end{align}
where $\oplus$ is the concatenation operation and $\odot$ is the element-wise product operation. Finally, we use a two-layer non-linear projection to obtain the entity label distribution for every span $s_{i,j}$:
\begin{align}
\textbf{r}_{i,j} &= tanh(W\textbf{s}_{i,j}) \\
\textbf{o}_{i,j} &= softmax(V\textbf{r}_{i,j})
\end{align}
where $W$ and $V$ are trainable parameter matrices.

And the probability of $l$-th gold entity label for span instance $s_{i,j}$ is $\textbf{o}_{i,j,l}$:
\begin{align}
\textbf{o}_{i,j,l} &= \textbf{v}_{l}^{T}\textbf{r}_{i,j}
\end{align}

We use cross entropy (CE) loss as our training objective:
\begin{align}
loss_{ce} = \sum_{s_{i,j}\in D} -log(\textbf{o}_{i,j,l})
\end{align}
where $D$ is the collection of all training instances.


\subsection{Span-based Contrastive Learning}
\noindent
To mitigate the ambiguity among entities, SCL-RAI tries to pull span belonging to the same class together in embedding space, while simultaneously pushing apart clusters of span from different classes.

This way, the clusters in entity representation space could better distinguish different types of entities. 
To this end, we propose a novel span-based contrastive learning objective to mitigate the ambiguity problem among entities. 
Meanwhile, we find in our experiment that this contrastive learning objective could also improve the robustness of SCL-RAI under unlabeled entity noises.

For span-based NER model, we conduct contrastive learning within a batch of span instances $D$;
We use the cosine similarity to represent the distance between the span representations of two instances $s_{i,j}$ and $s_{\hat i, \hat j}$:
\begin{equation}
    d_{s_{i,j},s_{\hat i, \hat j}} = \frac{\textbf{r}_{i,j} \cdot \textbf{r}_{\hat i,\hat j}}{\lvert \textbf{r}_{i,j} \rvert \lvert \textbf{r}_{\hat i,\hat j} \rvert}
\end{equation}

Then the span-based supervised contrastive learning loss function $loss_{scl}$ is defined as:
\begin{gather}
\begin{small}
loss_{scl} = -\sum_{l \in L}\sum_{s_{i,j} \in D_{l}} \frac{1}{N_l-1} \sum_{s_{\hat i,\hat j} \in D_{l}} F(\textbf{r}_{i,j},\textbf{r}_{\hat i,\hat j})
\end{small}
\end{gather}
where $L$ is the size of the entity label set; $(i, j) \ne (\hat i, \hat j)$; $N_{l}$ is the total number of span instances with the same entity label $l$ in the batch; 
$D_{l}$ is the collection of all training span instance with $l$-th entity label.  $F(\textbf{r}_{i,j},\textbf{r}_{\hat i,\hat j})$ is:
\begin{align}
F(\textbf{r}_{i,j},\textbf{r}_{\hat i,\hat j}) = log \frac{exp(d_{s_{i,j},s_{\hat i,\hat j}} / \tau) }{\sum_{s_{m,n} \in D_{\bar{l}}} exp(d_{s_{i,j},s_{m,n}} / \tau)}
\end{align}
where $\tau$ is the temperature. $D_{\bar{l}}$ is the collection of span instances not with $l$-th entity label.

This span-based supervised contrastive learning loss pushes the span representations of instances with the same entity labels closer and pushes the span representations of instances with the different entity labels farther. 
We confirm in our experiment that this contrastive learning objective indeed improves the model robustness under unlabeled entities, compared with previous works. 

Then we combine the cross entropy loss and span-based contrastive learning loss to get our final loss function: 
\begin{align}
loss_{final} = (1-\lambda)*loss_{ce} + \lambda * loss_{scl} 
\end{align}
where $\lambda$ is a scalar hyperparameter.



\subsection{Retrieval Augmented Inference}
\label{RAI}
\noindent
As we get the discriminative entity span representations via span-based contrastive Learning, we propose \textbf{R}etrieval \textbf{A}ugmented \textbf{I}nference (\textbf{RAI}) to facilitate the decoding process at the inference stage. As shown in Figure \ref{Fig_new}, RAI can be split into two parts:
(i) Firstly, it generates a central point representation for each entity type from the training set and stores them in a dictionary $Dict$.
(ii) It calculates the similarity between the representation of span to be predicted and each entity type representation in $Dict$ to get the retrieval augmented label distribution $\textbf{o}_{RA}$, then interpolates the distribution $\textbf{o}_{i,j,l}$ from span-based NER model with $\textbf{o}_{RA}$ to get the final label distribution.
For example, the second ``NBA'' in Figure~\ref{fig_example} will get high similarity value with the the central point representation of the entity type ``ORG'', due to the similar context with other ``ORG'' entities in training set. So it could decrease the possible high probability of non-entity label from Span-based NER model and increase it of ``ORG'' entity label. This way, we can 
shift the learned decision boundary toward the expected boundary in Figure~\ref{decision boundary bias phenomenon}.


\noindent
\textbf{Dictionary Construction}:\quad The dictionary $Dict$ used in SCL-RAI consists of a set of key-value pairs. Each key is an entity type and the corresponding value is the calculated central point representation from the training set. After training the model, we could get the dictionary for storing representations of all entity tags:
\begin{align}
Dict &= \{K,V\} = \{(l, \textbf{r}_{l}) | \forall l \in L \} \\
\textbf{r}_{l} &= \sum_{s_{m,n} \in T_{l}} \frac{1}{N_{l}}\textbf{r}_{m,n}
\end{align}
where $T_{l}$ is the collection of all training span instances with $l$-th entity label; $N_{l}$ is the total number of span instances with the label $l$ in the training set.

\noindent
\textbf{Label Distribution Interpolation}:\quad At the same time, Span-based NER model outputs representation $\textbf{r}_{i,j}$ for the span to be predicted and its label distribution $\textbf{o}_{i,j}$. 
Then we calculate the cosine similarity between $\textbf{r}_{i,j}$ each cached representation from $Dict$ to obtain a new label distribution, i.e., retrieval augmented label distribution $\textbf{o}_{RA}$:
\begin{align}
\textbf{sim}_{(i,j)} &= concat(\frac{\textbf{r}_{i,j} \cdot \textbf{r}_{l_1}}{\lvert \textbf{r}_{i,j} \rvert \lvert \textbf{r}_{l_1} \rvert},...,\frac{\textbf{r}_{i,j} \cdot \textbf{r}_{l_L}}{\lvert \textbf{r}_{i,j} \rvert \lvert \textbf{r}_{l_L} \rvert}) \\
\textbf{o}_{RA} &= softmax(\textbf{sim}_{(i,j)})
\end{align}
where $L$ is the number of entity labels.

We then set the value of non-entity label in $\textbf{o}_{RA}$ to 0:
\begin{align}
\textbf{o}_{RA}[v] = 0
\end{align}
where $v$ is the index for the non-entity. This ensures the similarity of non-entity label does not participate in interpolation.

Finally, we interpolate the distribution $\textbf{o}_{i,j,l}$ from span-based NER model with $\textbf{o}_{RA}$ to get the final label distribution $\textbf{p}_{final}$:
\begin{align}
\textbf{p}_{final} = (1-\alpha) * \textbf{o}_{i,j} + \alpha * \textbf{o}_{RA}
\end{align}
where $\alpha$ is a hyperparameter to makes a balance between two distributions.

\section{Experiments}

\subsection{Experimental Settings}
\noindent
Following \cite{DBLP:conf/coling/YangCLHZ18,DBLP:conf/iclr/LiL021,li-etal-2022-rethinking}, we adopt EC and NEWS as our datasets. 
The training set of EC and NEWS both consist of two parts: (1) the well-annotated set $\mathcal{A}$; (2) the distantly supervised set $\mathcal{DS}$, which contains a large amount of incompletely annotated sentences.
Therefore, NER models trained on EC or NEWS suffer from \textit{Unlabeled Entity Problem}. The dev/test set used in two datasets are well-annotated to evaluate the performance of model trained on datasets containing label noise. 

\noindent
\textbf{EC} \quad  In the e-commerce domain (EC), there are five types of entities: Brand, Product, Model, Material, and
Specification.
It contains 2,400 sentences labeled by annotators. 
The well-annotated set $\mathcal{A}$ is split into three sets: 1,200 sentences for training, 400 for dev, and 800 for testing.
Then \citet{DBLP:conf/coling/YangCLHZ18} collect a list of entities to construct a dictionary from the training data and perform distant supervision on raw data to get the distantly supervised set $\mathcal{DS}$, which contains 2,500 sentences.

\noindent
\textbf{NEWS} \quad  
For news domain, \citet{DBLP:conf/coling/YangCLHZ18} use a NER data from MSRA \cite{levow-2006-third}.
\citet{DBLP:conf/coling/YangCLHZ18} only keep entity type PERSON to get NEWS.
Then \citep{DBLP:conf/coling/YangCLHZ18} randomly select 3,000 sentences as training dataset, 3,328 as dev data, and 3,186 as testing data to get the well-annotated set $\mathcal{A}$. 
The rest set of MSRA is used as raw data, having 36,602 sentences. 
\citet{DBLP:conf/coling/YangCLHZ18} collect a list of person names from the training data. 
Then \citet{DBLP:conf/coling/YangCLHZ18} add additional names to the list. 
Finally, the list has 71,664 entries. 
\citet{DBLP:conf/coling/YangCLHZ18} perform distant supervision on raw data to obtain extra 3,722 sentences as the distantly supervised set $\mathcal{DS}$.


We adopt the same hyperparameter configurations for two datasets. 
We use Adam \cite{DBLP:journals/corr/KingmaB14} as optimizer with learning rate as $10^{-5}$ and bert-base as our encoder following \citet{DBLP:conf/iclr/LiL021,li-etal-2022-rethinking}. 
The dimension of scoring layers $W$ is set as 256. 
The scalar weighting hyperparameters $\lambda$ and $\alpha$ are set as 0.1 and 0.5. The temperature parameter $\tau$ is set to 0.1. 
Since the label distribution is very unbalanced (most instances are non-entity), we also apply negative sampling and the same sampling rate as \citet{DBLP:conf/iclr/LiL021}.

For evaluation, we use conlleval script\footnote{https://www.clips.uantwerpen.be/conll2000/chunking
/conlleval.txt.} to compute the F1-score.

\begin{table}
\scriptsize
\centering
\setlength{\tabcolsep}{3.0mm}{
\renewcommand{\arraystretch}{1.1}
\begin{tabular}{lcc}
\hline
\textbf{Models}&{EC}&{NEWS}\\
\hline
{BERT-MRC} &{55.72} &{74.55}\\
{BERT-Biaffine Model} &{55.99} & {74.57}\\
\hdashline
{PU Learning} &{61.22} & {77.98}\\
{Partial CRF} & {60.08} & {78.38}\\
{Weighted Partial CRF} &{61.75} & {78.64}\\
{Vanilla Negative Sampling} &{66.17} & {85.39}\\
{Variant Negative Sampling} &{67.03} & {86.15}\\
\hline
{\textbf{SCL-RAI}} &{69.70} & {94.11} \\
\textbf{SCL-RAI}+Vanilla Neg. Sampl. &{\textbf{71.24}} & {\textbf{94.79}}\\
\quad  {- RAI} & {70.65 (-0.59)} & {93.71 (-1.08)} \\
\quad  {- SCL \& RAI} & {66.17 (-5.07)} & {85.39 (-9.40)} \\
\hline
\end{tabular}}
\caption{The F1-score results on two real-world datasets. ``SCL'' denotes Span-based Contrastive Learning and ``RAI'' denotes Retrieval Augmented Inference.}
\label{tab_1} 
\end{table}

\begin{table}
\small
\centering
\setlength{\tabcolsep}{1.4mm}{
\renewcommand{\arraystretch}{1.1}
\begin{tabular}{lccc}
\hline
\textbf{Models} &{$\mathcal{A}$} &{$\mathcal{A} + \mathcal{DS}$} & {$\Delta$}\\
\hline
{Vanilla Negative Sampling} & {94.38} & {85.39}& {-8.99}\\
{\textbf{SCL-RAI}+Vanilla Neg. Sampl.} &{\textbf{95.33}}&{\textbf{94.79}}&{\textbf{-0.54}}\\
\hline
\end{tabular}}
\caption{The robustness of SCL-RAI over unlabeled entities with different training set on NEWS dataset.}
\label{robustness_NEWS} 
\end{table}

\begin{figure}
    \centering
    \includegraphics[width=7cm]{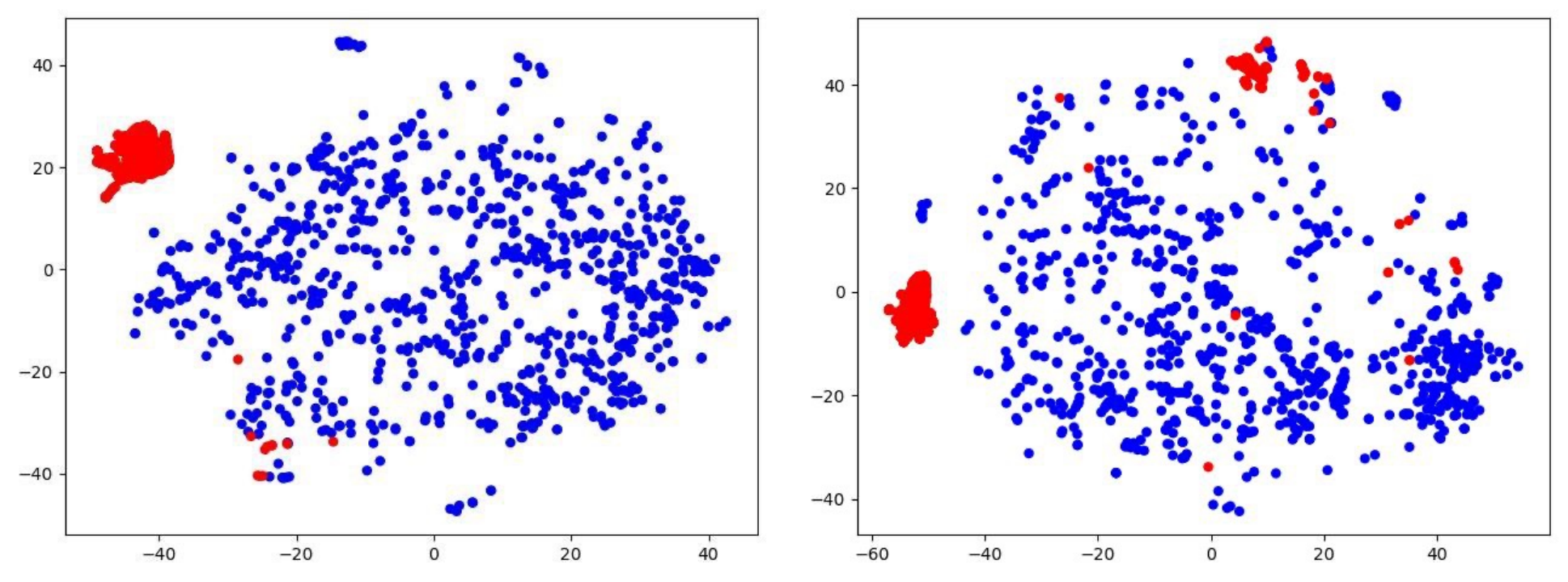}
    \caption{t-SNE plots of the representations on NEWS test set. CE+Span-based CL (left), CE only (right). Red dot denote entities and blue dot denote the non-entities.}
    \label{files:scatter}
\end{figure}

\subsection{Results and Analysis}
\noindent
We report the results from: 
(1) \textbf{Traditional NER methods}:
BERT-MRC \cite{yu-etal-2020-named} and BERT-Biaffine \cite{yu-etal-2020-named} ;
(2) \textbf{Recent Attempts on Unlabeled Entity Problem}:  PU Learning \cite{DBLP:conf/acl/PengXZFH19}, Partial CRF \cite{DBLP:conf/coling/YangCLHZ18}, 
Weighted Partial CRF \cite{jie-etal-2019-better}, Vanilla Negative Sampling \cite{DBLP:conf/iclr/LiL021}, Variant Negative Sampling \cite{li-etal-2022-rethinking} and our \textbf{SCL-RAI}. Since our method is orthogonal to that of \cite{DBLP:conf/iclr/LiL021}, we also report the results of SCL-RAI with their negative sampling strategy ``\textbf{SCL-RAI}+Vanilla Neg. Sampl.'' to get better results.

We report our results in Table \ref{tab_1}. Firstly, traditional NER models perform poorly on real-world datasets. 
So the SOTA NER models on well-annotated datasets are not robust over the \textit{Unlabeled Entity Problem}.
Then, our method has achieved new state-of-the-art results on the two datasets.
Compared with SOTA model \cite{li-etal-2022-rethinking}, we achieve the improvements of 2.67\% F1 on EC and 7.96\% on NEWS. With the negative sampling strategy, we further get the improvements of 4.21\% F1 on EC and 8.64\% F1 on NEWS. The improvements shows that our model has a stronger ability to mitigate the noise from unlabeled entities.

To verify the effectiveness of SCL-RAI, we show ablation studies in Table \ref{tab_1}. 
It is clear that Span-based Contrastive Learning and Retrieval Augmented Inference are both important to cope with \textit{Unlabeled Entity Problem}. In Table~\ref{robustness_NEWS} and Table~\ref{robustness_EC}, we show the robustness of our model over unlabeled entities on NEWS dataset. Our SCL-RAI can obtain less F1 degradation when introducing dataset $\mathcal{DS}$ with unlabeled entities. 
In Figure \ref{files:scatter}, we show t-SNE plots of the learned representations of 2000 instances on NEWS test set, comparing Cross Entropy (CE) with and without the Span-based CL term. 
We can clearly see that the Span-based CL term enforces more compact clustering of entities. 

For span-based NER model, we also conduct our Span-based Contrastive Learning within a batch of span instances.
Therefore, we test the robustness of Span-based Contrastive Learning for different batch sizes on EC.
As shown in Table \ref{tab_appendix}, we can clearly find that Span-based Contrastive Learning is robust for different batch sizes.

\begin{table}
\small
\centering
\setlength{\tabcolsep}{4mm}{
\renewcommand{\arraystretch}{1.1}
\begin{tabular}{cc}
\hline
{Batch Size} &{F1-score}\\
\hline
{8} & {70.78}\\
{16} & {71.24}\\
{32} & {70.87}\\
{64} & {70.79}\\
\hline
\end{tabular}}
\caption{The Span-based Contrastive Learning robustness testing results on EC.}
\label{tab_appendix} 
\end{table}

\begin{table}
\small
\centering
\setlength{\tabcolsep}{1.4mm}{
\renewcommand{\arraystretch}{1.1}
\begin{tabular}{cccc}
\hline
\textbf{Variant} &{$\mathcal{A}$} &{$\mathcal{A} + \mathcal{DS}$} & {$\Delta$}\\
\hline
{Vanilla Negative Sampling} & {76.82} & {66.17}& {-10.65}\\
{SCL-RAI+Vanilla Neg. Sampl.} &{\textbf{76.44}}&{\textbf{71.24}}&{\textbf{-5.2}}\\
\hline
\end{tabular}}
\caption{The robustness of SCL-RAI over unlabeled entities with different training set on EC dataset.}
\label{robustness_EC} 
\end{table}

\section{Conclusion}
\noindent
We propose the SCL-RAI to cope with \textit{Unlabeled Entity Problem} in NER. Benefiting from our well-designed Span-based Contrastive Learning and Retrieval Augmented Inference, experiments on two real-world datasets show that SCL-RAI achieves more promising results than SOTA methods.

\section{Acknowledgements}
\noindent
This paper is supported by the National Key R\&D Program of China under Grand No.2018AAA0102003, the National Science Foundation of China under Grant No.61936012 and 61876004. Our code is now available at \href{https://github.com/PKUnlp-icler/SCL-RAI}{https://github.com/PKUnlp-icler/SCL-RAI}.

\bibliography{anthology,custom}

\begin{thebibliography}{17}
\expandafter\ifx\csname natexlab\endcsname\relax\def\natexlab#1{#1}\fi

\bibitem[{Devlin et~al.(2019)Devlin, Chang, Lee, and
  Toutanova}]{DBLP:conf/naacl/DevlinCLT19}
Jacob Devlin, Ming{-}Wei Chang, Kenton Lee, and Kristina Toutanova. 2019.
\newblock \href {https://doi.org/10.18653/v1/n19-1423} {{BERT:} pre-training of
  deep bidirectional transformers for language understanding}.
\newblock In \emph{Proceedings of the 2019 Conference of the North American
  Chapter of the Association for Computational Linguistics: Human Language
  Technologies, {NAACL-HLT} 2019, Minneapolis, MN, USA, June 2-7, 2019, Volume
  1 (Long and Short Papers)}, pages 4171--4186. Association for Computational
  Linguistics.

\bibitem[{Fries et~al.(2017)Fries, Wu, Ratner, and
  R{\'{e}}}]{DBLP:journals/corr/Fries0RR17}
Jason~A. Fries, Sen Wu, Alexander Ratner, and Christopher R{\'{e}}. 2017.
\newblock \href {http://arxiv.org/abs/1704.06360} {Swellshark: {A} generative
  model for biomedical named entity recognition without labeled data}.
\newblock \emph{CoRR}, abs/1704.06360.

\bibitem[{Jie et~al.(2019)Jie, Xie, Lu, Ding, and Li}]{jie-etal-2019-better}
Zhanming Jie, Pengjun Xie, Wei Lu, Ruixue Ding, and Linlin Li. 2019.
\newblock \href {https://doi.org/10.18653/v1/N19-1079} {Better modeling of
  incomplete annotations for named entity recognition}.
\newblock In \emph{Proceedings of the 2019 Conference of the North {A}merican
  Chapter of the Association for Computational Linguistics: Human Language
  Technologies, Volume 1 (Long and Short Papers)}, pages 729--734, Minneapolis,
  Minnesota. Association for Computational Linguistics.

\bibitem[{Kingma and Ba(2015)}]{DBLP:journals/corr/KingmaB14}
Diederik~P. Kingma and Jimmy Ba. 2015.
\newblock \href {http://arxiv.org/abs/1412.6980} {Adam: {A} method for
  stochastic optimization}.
\newblock In \emph{3rd International Conference on Learning Representations,
  {ICLR} 2015, San Diego, CA, USA, May 7-9, 2015, Conference Track
  Proceedings}.

\bibitem[{Lafferty et~al.(2001)Lafferty, McCallum, and
  Pereira}]{DBLP:conf/icml/LaffertyMP01}
John~D. Lafferty, Andrew McCallum, and Fernando C.~N. Pereira. 2001.
\newblock \href
  {https://repository.upenn.edu/cgi/viewcontent.cgi?article=1162&context=cis_papers}
  {Conditional random fields: Probabilistic models for segmenting and labeling
  sequence data}.
\newblock In \emph{Proceedings of the Eighteenth International Conference on
  Machine Learning {(ICML} 2001), Williams College, Williamstown, MA, USA, June
  28 - July 1, 2001}, pages 282--289. Morgan Kaufmann.

\bibitem[{Levow(2006)}]{levow-2006-third}
Gina-Anne Levow. 2006.
\newblock \href {https://aclanthology.org/W06-0115} {The third international
  {C}hinese language processing bakeoff: Word segmentation and named entity
  recognition}.
\newblock In \emph{Proceedings of the Fifth {SIGHAN} Workshop on {C}hinese
  Language Processing}, pages 108--117, Sydney, Australia. Association for
  Computational Linguistics.

\bibitem[{Li et~al.(2022{\natexlab{a}})Li, Li, Feng, and Ouyang}]{li2022who}
Changchun Li, Ximing Li, Lei Feng, and Jihong Ouyang. 2022{\natexlab{a}}.
\newblock \href {https://openreview.net/forum?id=NH29920YEmj} {Who is your
  right mixup partner in positive and unlabeled learning}.
\newblock In \emph{International Conference on Learning Representations}.

\bibitem[{Li et~al.(2022{\natexlab{b}})Li, Fei, Liu, Wu, Zhang, Teng, Ji, and
  Li}]{li2022unified}
Jingye Li, Hao Fei, Jiang Liu, Shengqiong Wu, Meishan Zhang, Chong Teng,
  Donghong Ji, and Fei Li. 2022{\natexlab{b}}.
\newblock \href {https://www.aaai.org/AAAI22Papers/AAAI-742.LiJ.pdf} {Unified
  named entity recognition as word-word relation classification}.
\newblock In \emph{Proceedings of the AAAI Conference on Artificial
  Intelligence}, volume~36, pages 10965--10973.

\bibitem[{Li and Liu(2005)}]{DBLP:conf/ecml/LiL05}
Xiaoli Li and Bing Liu. 2005.
\newblock \href {https://doi.org/10.1007/11564096\_24} {Learning from positive
  and unlabeled examples with different data distributions}.
\newblock In \emph{Machine Learning: {ECML} 2005, 16th European Conference on
  Machine Learning, Porto, Portugal, October 3-7, 2005, Proceedings}, volume
  3720 of \emph{Lecture Notes in Computer Science}, pages 218--229. Springer.

\bibitem[{Li et~al.(2021)Li, Liu, and Shi}]{DBLP:conf/iclr/LiL021}
Yangming Li, Lemao Liu, and Shuming Shi. 2021.
\newblock \href {https://openreview.net/forum?id=5jRVa89sZk} {Empirical
  analysis of unlabeled entity problem in named entity recognition}.
\newblock In \emph{9th International Conference on Learning Representations,
  {ICLR} 2021, Virtual Event, Austria, May 3-7, 2021}. OpenReview.net.

\bibitem[{Li et~al.(2022{\natexlab{c}})Li, Liu, and
  Shi}]{li-etal-2022-rethinking}
Yangming Li, Lemao Liu, and Shuming Shi. 2022{\natexlab{c}}.
\newblock \href {https://doi.org/10.18653/v1/2022.acl-long.497} {Rethinking
  negative sampling for handling missing entity annotations}.
\newblock In \emph{Proceedings of the 60th Annual Meeting of the Association
  for Computational Linguistics (Volume 1: Long Papers)}, pages 7188--7197,
  Dublin, Ireland. Association for Computational Linguistics.

\bibitem[{Ling and Weld(2012)}]{DBLP:conf/aaai/LingW12}
Xiao Ling and Daniel~S. Weld. 2012.
\newblock \href {http://www.aaai.org/ocs/index.php/AAAI/AAAI12/paper/view/5152}
  {Fine-grained entity recognition}.
\newblock In \emph{Proceedings of the Twenty-Sixth {AAAI} Conference on
  Artificial Intelligence, July 22-26, 2012, Toronto, Ontario, Canada}. {AAAI}
  Press.

\bibitem[{Peng et~al.(2019)Peng, Xing, Zhang, Fu, and
  Huang}]{DBLP:conf/acl/PengXZFH19}
Minlong Peng, Xiaoyu Xing, Qi~Zhang, Jinlan Fu, and Xuanjing Huang. 2019.
\newblock \href {https://doi.org/10.18653/v1/p19-1231} {Distantly supervised
  named entity recognition using positive-unlabeled learning}.
\newblock In \emph{Proceedings of the 57th Conference of the Association for
  Computational Linguistics, {ACL} 2019, Florence, Italy, July 28- August 2,
  2019, Volume 1: Long Papers}, pages 2409--2419. Association for Computational
  Linguistics.

\bibitem[{Ren et~al.(2015)Ren, El{-}Kishky, Wang, Tao, Voss, and
  Han}]{DBLP:conf/kdd/RenEWTVH15}
Xiang Ren, Ahmed El{-}Kishky, Chi Wang, Fangbo Tao, Clare~R. Voss, and Jiawei
  Han. 2015.
\newblock \href {https://doi.org/10.1145/2783258.2783362} {Clustype: Effective
  entity recognition and typing by relation phrase-based clustering}.
\newblock In \emph{Proceedings of the 21th {ACM} {SIGKDD} International
  Conference on Knowledge Discovery and Data Mining, Sydney, NSW, Australia,
  August 10-13, 2015}, pages 995--1004. {ACM}.

\bibitem[{Shang et~al.(2018)Shang, Liu, Jiang, Ren, Voss, and
  Han}]{DBLP:journals/tkde/ShangLJRVH18}
Jingbo Shang, Jialu Liu, Meng Jiang, Xiang Ren, Clare~R. Voss, and Jiawei Han.
  2018.
\newblock \href {https://doi.org/10.1109/TKDE.2018.2812203} {Automated phrase
  mining from massive text corpora}.
\newblock \emph{{IEEE} Trans. Knowl. Data Eng.}, 30(10):1825--1837.

\bibitem[{Yang et~al.(2018)Yang, Chen, Li, He, and
  Zhang}]{DBLP:conf/coling/YangCLHZ18}
YaoSheng Yang, Wenliang Chen, Zhenghua Li, Zhengqiu He, and Min Zhang. 2018.
\newblock \href {https://aclanthology.org/C18-1183/} {Distantly supervised
  {NER} with partial annotation learning and reinforcement learning}.
\newblock In \emph{Proceedings of the 27th International Conference on
  Computational Linguistics, {COLING} 2018, Santa Fe, New Mexico, USA, August
  20-26, 2018}, pages 2159--2169. Association for Computational Linguistics.

\bibitem[{Yu et~al.(2020)Yu, Bohnet, and Poesio}]{yu-etal-2020-named}
Juntao Yu, Bernd Bohnet, and Massimo Poesio. 2020.
\newblock \href {https://doi.org/10.18653/v1/2020.acl-main.577} {Named entity
  recognition as dependency parsing}.
\newblock In \emph{Proceedings of the 58th Annual Meeting of the Association
  for Computational Linguistics}, pages 6470--6476, Online. Association for
  Computational Linguistics.

\end{thebibliography}
\bibliographystyle{acl_natbib}



\clearpage
\end{document}